\documentclass[twocolumn]{article}
\usepackage{times,amsmath,epsfig,multirow,hyperref}
\title{{\huge A Survey on Recent Advances of Computer Vision Algorithms for Egocentric Video}}
\author{
{Sven Bambach}
\vspace{1.6mm}\\
\fontsize{10}{10}\selectfont\itshape
School of Informatics and Computing\\
\fontsize{10}{10}\selectfont\itshape
Indiana University\\
\fontsize{10}{10}\selectfont\itshape
Bloomington, IN
\vspace{1.6mm}\\
\fontsize{9}{9}\selectfont\ttfamily\upshape
\,sbambach@indiana.edu
}
\date{September 8, 2013}
\begin{document}

\maketitle
\begin{abstract} 
\textit{Recent technological advances have made lightweight, head mounted cameras both practical and affordable and products like Google Glass show first approaches to introduce the idea of egocentric (first-person) video to the mainstream. Interestingly, the computer vision community has only recently started to explore this new domain of egocentric vision, where research can roughly be categorized into three areas: Object recognition, activity detection/recognition, video summarization. In this paper, we try to give a broad overview about the different problems that have been addressed and collect and compare evaluation results. Moreover, along with the emergence of this new domain came the introduction of numerous new and versatile benchmark datasets, which we summarize and compare as well.}
\end{abstract}

\section{Introduction}

Most of the classic work in computer vision has been devoted to studying either static images or video from stationary cameras (such as tracking objects in surveillance applications). Recently, technological advances have made lightweight, wearable, egocentric cameras both practical and popular in various fields. The \textit{GoPro} camera for instance can be mounted to helmets and is popular in a lot of sports such as biking, surfing or skiing. The \textit{Microsoft SenseCam} can be worn around the neck and has enough video storage to capture an entire day for the idea of ``life logging''. Cognitive scientists like to use first-person cameras attached to glasses (often in combination with eye trackers such as \textit{Tobii} or \textit{SMI}) to study visual attention in naturalistic environments. Most recently, emerging products like \textit{Google Glass} started making first attempts to bring the idea of wearable, egocentric cameras into the mainstream.

From a computer vision standpoint, videos from these first-person devices pose a lot of challenges. Because the camera is constantly moving, the motion is highly nonlinear and unpredictable. As a result, objects may rapidly disappear and reappear in the field of view. In extreme cases (such as sport videos), one must also expect things like motion blur, splashing water or glare. On the other hand, some qualities of egocentric video may be helpful for specific applications. For example, objects that the observer manipulates or people and faces that the observer interacts with, tend to naturally be centered in the view and are less likely to be occluded then they might be if captured from a static, third person camera.

In the next section, we will introduce the most recent work from the computer vision community in the domain of egocentric video. We further try to point out egocentric-specific challenges that occurred within the given problems, but also mention situations were the egocentric paradigm was actually useful. We emphasize that egocentric video is an emerging field and a lot of the work that we will reference can be considered as pioneering work. As a result of that, not many things are built on top of each other and direct quantitative comparisons between different works are often difficult. 

Another effect of the novel nature of the egocentric domain is the emergence of numerous new and very versatile data sets. While briefly explaining the individual data sets along with the work in section \ref{sec:work}, we give a detailed overview about publicly available datasets in section \ref{sec:data}. 

In section \ref{sec:summary}, we summarize and compare results from the previous sections and finally section \ref{sec:conclusion} concludes the paper.

\section{Recent Work}
\label{sec:work}

In this section, we introduce recent work in the field of egocentric video. We group this work into three categories. The first category deals with object recognition with respect to objects that are being manipulated (by hand) by the first-person observer. The second category deals with the detection and recognition of first-person actions and activities. We will see that this category naturally emerges from the first one, as most of the considered activities are characterized by the objects being used. The third category deals with so called ``life logging'' video data. This data is mainly characterized by the fact that it involves hourlong, continuous video data depicting the ``life'' of the first-person observer. Work in this area usually deals with data summarization, i.e. the extraction of relevant or representative frames or actions. However, there is also work in more specific tasks such as the detection of social interactions based on egocentric video recorded by a group of  people in a theme park.

\subsection{Object Recognition}

One of the first analyses of object recognition in egocentric video was done by Ren and Philipose \cite{ren2009egocentric}. Motivated by the idea that recognizing handled objects can provide essential information about a person's activity, they wanted to explore the challenges and characteristics of object recognition in the context of egocentric video. They collected a video dataset consisting of 42 everyday objects (milk carton, watering can, etc.), where each object was being manipulated by hands in an object-specfic way. To obtain some baseline results for their dataset, they annotated a small subset of frames with ground-truth object versus background segmentations. They used a standard SIFT based recognition system described in \cite{lowe1999object} and trained a multi-class SVM. They achieved a 12\% recognition rate compared to a random chance of 2.4\%. They went on to quantify the influence of various egocentric-specific challenges, such as limited texture of objects, background clutter and hand occlusion. To gain an upper bound for recognition performance, they used the SIFT recognition system on clean exemplar images of their objects, obtaining an average accuracy of 63.7\%. Simulating occlusion on the clean exemplars had the accuracy drop down to 57.0\% while simulating background clutter resulted in a 20\% drop in accuracy and combining both had the accuracy drop down to 30.3\%. They suggest motion and location priors as well as hand detection as future research directions.

Follow-up work has been done by Ren and Gu \cite{ren2010figure} who developed a motion-based approach to segment out foreground objects in egocentric video in order to improve object recognition accuracy. The idea is based on the observation that there are some regularities with respect to motion in egocentric video that are useful towards motion segmentation: During object manipulation, hands and objects have the tendency to appear near the center of the view and body (i.e. camera) motions are rather small and horizontal. Their model explicitly addresses this with a motion prior and a location prior for each pixel. The distribution for the location prior is built by averaging ground truth segmentation masks and the motion prior is based on optical-flow results obtained from video parts that only contain background (no hands or objects), thus giving an average flow estimation for each background pixel. Additionally, they used temporal cues that take segmentation masks from previous frames into account. Finally, they used the coarse-to-fine variational optical flow algorithm of \cite{brox2004high} to create dense optical flow across two frames and then used RANSAC to fit the motion vectors into affine layers. Equipped with these motion features and priors, they trained a max-margin classifier for pixelwise figure-ground classification and cleaned up the results using the standard Graph Cut algorithm. For testing, they used the same 42 object dataset as \cite{ren2009egocentric} and improved the accuracy of the SIFT based recognition system from 12\% to 20\%. They also tested a latent HOG based recognition system \cite{felzenszwalb2008discriminatively} and found that the accuracy improved from 38\% to 46\%.

Fathi \textit{et al.} \cite{fathi2011learning} took advantage of the egocentric paradigm (objects of interest tend to be centered and at a large scale) to learn object classification and segmentation with very weak supervision. The motivating idea to use object recognition as a way to make inference on possible activities is similar to that of \cite{ren2009egocentric}, but is taken a step further in the sense that they explored egocentric activities involving multiple objects (such as making a peanut butter and jelly sandwich). They hypothesized that the co-occurence of different objects within those activities can be exploited for object detection and localization. They performed figure-ground segmentation as well, but their approach differed from \cite{ren2010figure} as it allowed objects to become part of the background after being manipulated. This is accomplished by splitting the video into short intervals and creating a local background model for each. For the weakly supervised learning, they collected a dataset of 7 daily activities involving multiple objects (making coffee/tee/sandwiches). Each video was only labeled with the list of objects it contained. To learn an appearance model for each object type, they used the diverse density based multiple instance learning framework of \cite{chen2006miles}. They further used equality constraints to assign the same label to regions with significant temporal connections. The object recognition accuracy ranged from about 10\% (sugar) to about 95\% (coffee). Additionally, their figure-background segmentation approach outperformed \cite{ren2010figure} on the 42 object dataset, having a 48\% segmentation error rate as opposed to 67\%.

\subsection{Activity and Action Detection}

Many authors recognized that a lot of activities that are interesting from an egocentric perspective are characterized by the observer manipulating objects in front of him. This is very different from third person videos where objects might be hard to see and thus, people focussed on activities that can be modeled by different body movements (e.g. dancing). In this section, we will use the terminology that has been established in recent work on egocentric activity and action detection, which is that actions describe simple, straightforward things such as ``take the knife'' or ``open the fridge'', while an activity describes a more complex aggregation of actions such as making coffee.

\subsubsection{Early Work Using Gist}

Early work in the domain of both unsupervised action segmentation and supervised action classification was done by Spriggs \textit{et al.} \cite{spriggs2009temporal}. They introduced the ``CMU kitchen'' dataset that contains multimodal measures, including egocentric video, of people cooking different recipes (brownies, pizza, etc.) in a kitchen environment. Each frame was labeled with an action class (such as ``stirring''). For action segmentation, rather than trying to recognize objects like most of the follow-up work, they computed the gist \cite{oliva2001modeling} of each frame. The assumption is that, under the egocentric paradigm, specific actions are performed in front of a somewhat constant background, making a gist feature vector a reasonable approach to model each frame. They performed PCA to reduce the vector dimensionality and estimated different Gaussian mixture models to investigate whether these features cluster into similar scenes. For some activities, such as ``stirring'', they saw promising results (70\% of frames labeled with this action were assigned to the right cluster) but noted that results do not generalize well as model parameters need to be varied to capture distinct sets of actions. They also explored supervised action classification by training an HMM with a mixture of Gaussians output on the gist features and obtained an average classification accuracy of 9.38\% (chance being 3\%). Lastly, they applied a simple KNN model, where each test frame from one subject is given the label of the frame with the smallest Euclidian distance from the set of frames of all other subjects, reaching a classification accuracy of 48.64\%.

\subsubsection{Object-based Activity Detection}

Further research on activity detection was done by Pirisiavash and Ramanan \cite{pirsiavash2012detecting}, whose work stands out due to their large, versatile and fully labeled dataset. They captured 18 daily indoor activities such as brushing teeth, washing dishes, or watching television, each performed by 20 different subjects in their respective apartments. 42 different object classes involved in these activities were annotated with bounding boxes. Each object also had a label depicting whether it is currently active (in hands) or not. Also driven by the idea that activities are all about the objects being involved, they used their data to build an activity model that explicitly models object use over time. For every frame of a given activity, they used the part-based object model by \cite{felzenszwalb2010object} to record a score based on the most likely position and scale for each of their 42 object classes. Averaging this score over all activity frames yielded a histogram of object scores for a specific activity. They went on to temporally split the video into halves in a pyramid fashion, each time calculating the object score histogram, and thus ending up with an activity model that describes object use over time. They learned a linear SVM on these models. Trained with all objects, they achieved a 32.6\% activity classification accuracy (chance being 5.6\%) and trained with only active objects they achieved 40.6\% accuracy.

An alternative, unsupervised activity model was proposed by Fathi \textit{et al.} \cite{fathi2011understanding}. Continuing their own work on object recognition in egocentric video \cite{fathi2011learning}, they proposed a graph based model that takes advantage of the semantic relationship between activities, actions and objects. They worked on the same dataset as they did in \cite{fathi2011learning}, which contains activities such as making various kinds of sandwiches. Based on detected objects, object-hand interactions and a set of action labels (``spread butter on bread'', etc.) they used an approach similar to Expectation-Conditional Maximization \cite{meng1993maximum} to learn actions and then learn activities from actions. Then, the inferred activity label was fixed and used to enhance action recognition results, as the activity can limit the set of possible actions as well as enforce a certain order. Finally, they enhanced their initial object recognition by learning a probabilistic object model that incorporates the inferred action priors. They recognized 6 out of 7 activities correctly and their action recognition accuracy was at 32.4\% (chance being 1.6\%). They also showed that this framework indeed improved their initial object recognition performance, achieving better results for almost all object classes.

Fathi \textit{et al.} extended their work in \cite{fathi2012learning} by additionally considering eye gaze, using calibrated, head-mounted eye trackers in combination with egocentric cameras. They raised the question whether knowing the fixation locations helps to better recognize actions and vice versa. This approach is motivated by psychological studies \cite{land2001ways} which demonstrate that during object manipulation tasks a substantial percentage of gaze fixations fall upon task-relevant objects. They used a generative model to describe the relationship between egocentric action and gaze location. This means they learned the probability of transitioning to a gaze location $g_t$, given $g_{t-1}$ and the current action $a$, as well as the likelihood of an image feature $x_t$, given the current action $a$ and the gaze position $g_t$. The image features were based on object features described in their earlier work \cite{fathi2011understanding}, as well as appearance features and future manipulation features. The appearance features were used to describe the fixated part of an object and were based on color and texture histograms in a circular area around the gaze location. Future manipulation features were aimed to take advantage of the fact that gaze is usually a split second ahead of the hands, so knowing the hand location a few frames ahead provides a cue of the gaze location in the current frame. They used a new dataset involving different kinds of meal preparations similar to their previous work \cite{fathi2011learning} but extended by the gaze data. They found that incorporating gaze information improved the action recognition accuracy to 47\% compared to 27\% when using the method of \cite{fathi2011understanding}. They also found promising results when predicting gaze locations given the action. However, when inferencing both action and gaze location action recognition accuracy only improves marginally (29\%).

\subsubsection{State-based Activity Detection}

Very recently, Fathi \textit{et al.} proposed a new approach to model actions in egocentric videos \cite{Fathi_2013_CVPR}, exploiting the fact that goal-oriented actions (``open coffee jar'') within object-manipulation activities (making coffee/sandwiches) can be detected by state changes of the objects being involved. Thus, for training purposes, they annotated each action with start frame, end frame, action label as well as a set of nouns describing the objects being involved. Focussing only on foreground objects \cite{fathi2011learning}, they discovered regions that changed before and after the action and clustered them into regions that constantly appear during the action to prune out irrelevant regions (such as hands). They then described those regions with color, texture and shape features and trained a linear SVM to learn a state-specific region detector. The action itself was then described as a quantized response of start and end frame to each region detector. With those responses, a second linear SVM was trained to build an action detector. They validated their model in terms of action recognition and activity segmentation, achieving a 39.7\% action recognition accuracy (based on 61 action classes) and outperformed their previous work in \cite{fathi2011understanding}. They achieved a 33\% accuracy for activity segmentation, based on the percentage of test video frames that had been labeled with the correct action.

\subsubsection{Interaction and Sport Activities}

Ryoo and Matthies recently were the first to explore interaction-level human activities from a first-person view \cite{S._2013_CVPR}. Motivated by surveillance, military or general human-robot interaction scenarios, they constructed a dataset of humans directly interacting with the egocentric observer. Interactions varied from friendly (shaking hands or petting the observer) to hostile (punching the observer or throwing objects at the observer). Based on the idea that interaction with the observer causes a lot of ego-motion, they used a combination of global and local motion descriptors to depict different activities. For global motion, they applied a conventional pixel-wise optical flow algorithm and built a histogram based on location and directions of the flow. For local motion, they interpreted the video as a 3-D XYT volume by concatenating frames over time and applied the cuboid feature detector by \cite{dollar2005behavior} to obtain video patches that contain salient motion. These motion descriptors were clustered using k-means to obtain a set of visual words. They represented an activity video as a histogram of these words and finally trained an SVM. Results were evaluated in terms of activity classification and detection, receiving a 89.6\% classification accuracy (based on 7 different activities), as well as an average detection precision of 0.709.

Kitani \textit{et al.} \cite{kitani2011fast} observed the increased usage of egocentric cameras in sport videos (biking, skiing, etc.). They developed a fast, unsupervised approach to index videos into different ego-actions that is supposed to help the athlete to locate and review specific parts without the burden of manual search. Similar to \cite{S._2013_CVPR}, they leveraged the fact that first-person sport videos contain lots of ego-motion and used optical flow histograms to describe the motions of a specific sport video. As a lot of the sport activities contain periodic movements, they additionally performed a DFT on the optical flow amplitudes to obtain frequency histograms. They used a Dirichlet mixture model \cite{wang2011fast} to first infer a motion codebook and then infer ego-action categories. They evaluated their performance on both controlled, choreographed videos as well real-world sport videos obtained from \textit{YouTube} and reported an F-measure (considering both precision and recall) for each sport. They achieved an F-measure of 0.93 for the choreographed videos and and average F-measure of 0.6 for the sport videos. Ego-actions varied between sports and involved labels such as ``hop down'', ``turn left'' or ``wedge left'' for skiing.

\subsection{Life Logging Video}

Another area that is particularly of interest in the ubiquitous computing community and contains egocentric video is the idea of ``life logging''. Here, a first-person camera continuously records a whole day of its wearer's life. The overall motivation that is mentioned by a lot of authors is to eventually develop systems that can serve as a retrospective memory aid for people with memory loss problems \cite{hodges2006sensecam}. Thus, a common goal is to summarize long, egocentric video or detect novel, anomalous events.

\subsubsection{Video Summarization}

Doherty \textit{et al.} \cite{doherty2008investigating} were among the first to investigate keyframe selection methods in the egocentric domain by looking at the \textit{Microsoft SenseCam}, a camera worn around the neck that takes an image every couple of seconds (an average of 1,900 images a day) to create a passively captured, visual life log. They pointed out that a lot of the established mechanisms for keyframe selection do not translate directly to the domain of life logging video, as they, for instance, rely on motion analysis and, due to the very low frame rate of their camera, motion is virtually non-existing. Also, passive capture devices may not always capture high quality images and hands or clothing covering parts of the lens are quite common. First, the authors split the set of images into different events where event boundaries are determined by high dissimilarity between frames according to a distance metric based on color and edge descriptors. They compared and investigated various approaches to select a keyframe for each of those events. Approaches varied from very simple solutions such as taking the middle image of the event, over taking the image that is closest to the average value of all images in the event, to more complex solutions like the image that is closest to the event average, farthest from the average of other events and performs well on various image quality tests for sharpness and contrast. Over 13,000 keyframes were judged by user ratings, where the most complex approach had a 8.4\% higher score than the base line (middle frame). They found that issues mainly occur during events that include a lot of motion (such as walking home) as there may be vast differences between images of the same event due to the nature of the camera and its low frame rate. 

Lee \textit{et al.} devised a method that aims to summarize life logging video material and goes beyond common keyframe detection by focussing on ``importance cues'' specific to the egocentric domain, such as objects and people the camera wearer interacts with \cite{lee2012discovering}. In particular, they segment each frame into multiple regions using a constrained parametric min-cuts method \cite{carreira2010constrained} and learn a regressor that predicts an importance score for each region. The score is based on a combination of various features: interaction (euclidean distance of region centroid to hand centroid, where hand is detected based on skin color), gaze (euclidean distance to center), frequency (appearance of region over multiple frames based on DoG+SIFT descriptors), object-like appearance (based on a ranking function of \cite{carreira2010constrained}), object-like motion, and likelihood of a person's face within a region (using the Viola-Jones method \cite{viola2001rapid}). They ended up temporally clustering the video into different events based on color histogram differences and represented each event with the frame that has the highest importance score based on the regressor. For training and evaluation, they used Amazon's \textit{Mechanical Turk} to manually label and segment important regions in their video data, which consisted of multiple hours of daily life activities among four different subjects. They evaluated the performance on classifying important regions correctly (by thresholding the regressor), as well as the quality of the keyframe summary. They found that their method performed better in predicting important objects than object-like features alone or low-level saliency methods. To quantify the perceived quality of the keyframe summaries, they asked the subjects that wore the camera to compare their method with baseline methods (such as uniform sampling among events), finding that their method was found better 68.75\% of the time.

Lu and Grauman \cite{Lu_2013_CVPR} extended this work by developing a story-driven (rather than object-driven) approach to summarize egocentric life logging video. The idea is to devise an influence metric that captures event connectivity and accounts for how one event leads to another, in order to create a summary that provides a better sense of a story. They also introduced a novel temporal segmentation method to cluster the video material into different events, which was specifically designed for egocentric video. They found that the method based on changes in color histograms which they used in previous work \cite{lee2012discovering} does not really work well for egocentric video due to its continuous nature. Instead, they tried to distinguish whether the camera wearer is static, in transit (physically traveling from one point to another), or moving the head. They learned an SVM to predict these scenarios based on dense optical flow features and blurriness scores \cite{crete2007blur}. They found that this method produced events (e.g. sets of frames) of an average length of 15 seconds. They represented each event in terms of detected objects. For known environments, objects were represented as scores based on a bank of object detectors and for uncontrolled environments, objects were essentially visual words based on object-like windows \cite{alexe2010object}. They went on to consider each event as a node in a chain. Finding a story-driven summary consisting of $k$ frames then comes down to finding the optimal, order-preserving $K$-node subchain with respect to story, importance and diversity constraints. Basically, the importance score was estimated similarly to their previous work \cite{lee2012discovering}, the story constraint favored event pairs with similar object instances, and the diversity constraint made sure that sequential events are not too similar. They found a good chain with the approximate best-first search strategy described in \cite{shahaf2010connecting}. They evaluated their performance in the form of a user study based on their own dataset \cite{lee2012discovering} as well as the ``Activities of Daily Living'' dataset from \cite{pirsiavash2012detecting}. To do so, they had 34 subjects compare their approach with other techniques such as uniform sampling or their previous work \cite{lee2012discovering}. They found that an average of 87\% of the subjects preferred their approach among different datasets and baselines.

\subsubsection{Novelty Detection}

Aghazadeh \textit{et al.} \cite{aghazadeh2011novelty} looked at videos from a subject who recorded his one-hour commute to work multiple times, wearing an egocentric camera that captures one image per second. Motivated by  the idea to use life logging cameras as a memory support system for the disabled \cite{hodges2006sensecam}, they proposed a method of novelty detection, where a novel event might be ``meeting a friend'' during the otherwise similar sequences of the subject going to work. They achieved this by exploiting the invariant temporal order of the activities across the different sequences to automatically align a query sequence with the other sequences. The idea is that a bad alignment yields a novelty in the query action as it is likely caused by an event that has not been observed in the reference sequences. They derived a similarity measure between two frames based on VLAD (vector of locally aggregated descriptors, proposed by \cite{jegou2010aggregating}) as well as geometric similarities, represented by the epipolar geometry between the two frames (i.e. the fundamental matrix). Comparing each frame from the query sequence with each frame from a reference sequence creates a cost matrix whose minimum cost path connecting the first and last frame (with the constraint that matches have to occur in temporal order) yields the best alignment between the two sequences. Finally, if a frame from the query sequence has a minimum match cost among all reference sequences that is above some threshold, it is considered a novelty. From 31 sequences of the subject going to work, four of them contained an event that the authors considered novel and all of them were detected by the algorithm.

\subsubsection{Social Interactions}

Fathi \textit{et al.} \cite{fathi2012social} looked at egocentric life logging video for social events, in particular people spending a day at an amusement park, and developed a method for the detection and recognition of social interactions. This was motivated by the idea that typically, one or more individuals have to play the role of the ``group videographer'' to capture memorable events, which prevents them from fully participating in the group experience. Moreover, a lot of memorable moments may occur spontaneously and the authors' thesis is that the presence or absence of social interactions is an important cue as to whether an event is viewed as memorable. The idea is that different kinds of social interactions can be detected/recognized by faces and their spatial attention. For instance, a monologue should have multiple observing faces attending the talking face. To model this, they first computed the orientation of each detected face using the Pittpatt face detection software\footnote{Pittpatt has since been acquired by Google Inc. and the software is not publicly available anymore.} and then used the camera's intrinsic parameters, as well as prior knowledge of face sizes at certain distances in order to estimate face locations and orientations in 3D. To get an estimate of the locations that the faces are attending, they built an MRF that incorporates these 3D locations/orientations as unary potentials, but also uses pairwise potentials between faces that bias nearby faces towards looking at the same location in the scene. They used an $\alpha$-expansion method to optimize the MRF. Having an estimate for each face's attention, they assigned roles to individual faces based on features such as the number of faces looking at $x$. Based on those, they could classify an interaction as dialogue, discussion, monologue and other labels, using a Hidden Conditional Random Field \cite{gunawardana2005hidden} that also incorporated temporal information. They reported results for both attention estimation as well as social interaction detection and recognition. Based on about 1000 hand-labeled frames, their method correctly estimated who is looking at whom in 71.4\% of the cases. For detection, they presented ROC curves for different forms of interaction, where the average area under the curve is 0.88. The average recognition accuracy was 55\% (chance being 20\%).

\section{Datasets}

\label{sec:data}

\begin{figure*}[]
{\footnotesize 

\begin{tabular}{| p{2.6cm} | p{4.1cm} | p{4.1cm}  | p{1.2cm} | p{2.8cm} |}
	\hline
	\textbf{Name} & \textbf{Description} & \textbf{Labeling} & \textbf{Used in} & \textbf{URL} \\
	\hline
	
	Intel 42 Objects & 10 video sequences (100K frames) from two human subjects manipulating 42 everyday object instances such as coffeepots, sponges, or cameras & each frame labeled with name of object; exemplar photos of objects with forground/background segmentation & \cite{ren2009egocentric, ren2010figure, fathi2011learning} & \url{http://seattle.intel-research.net/~xren/egovision09/} \\
	\hline
	\hline
	GeorgiaTech Egocentric Activities (GTEA)& 7 types of daily activities such as making a sandwhich/coffee/tea; each performed by 4 different subjects& each activity video is labeled with list of objects being involved; each frame has left hand, right hand, and background segmentation masks& \cite{fathi2011learning, fathi2011understanding, Fathi_2013_CVPR} & \url{http://www.cc.gatech.edu/~afathi3/GTEA/} \\
	\hline
	
	CMU kitchen & multimodal dataset of 18 subjects cooking 5 different recipes (brownies, pizza, etc.); also contains audio, body motion capture, and IMU data & each frame is labeled with an action such as ``take oil'', ``crack egg'', etc. & \cite{spriggs2009temporal} & \url{http://kitchen.cs.cmu.edu/} \\
	\hline
	
	Activities of Daily Living & 18 daily indoor activities such as brushing teeth, washing dishes, or watching television, each performed by 20 different subjects & 42 object classes that are involved in the activities are annotated with bounding boxes in all frames & \cite{pirsiavash2012detecting, Lu_2013_CVPR} & \url{http://deepthought.ics.uci.edu/ADLdataset/adl.html} \\
	\hline
	
	GeorgiaTech Egocentric Activities - Gaze+ & 7 types of meal preparation such as making pizza/pasta/salad; each performed by 5 different subjects   &each frame has eye gaze fixation data, timeframes of different activities such as ``open fridge'' are annotated & \cite{fathi2012learning} & \url{http://www.cc.gatech.edu/~afathi3/GTEA_Gaze_Website/} \\
	
	\hline
	\hline
	
	UT Egocentric & 4 videos from head-mounted cameras capturing a person's day, each about 3-5 hours long & not available & \cite{lee2012discovering,Lu_2013_CVPR} & \url{http://vision.cs.utexas.edu/projects/egocentric/index.html} \\
	\hline
	
	First-Person Social Interactions & day-long videos of 8 subjects spending their day at Disney World   & timeframes for different activities (``waiting'', ``train ride'', etc.) and social interactions (dialogue, discussion, etc.) are annotated & \cite{fathi2012social} & \url{http://www.cc.gatech.edu/~afathi3/Disney/} \\
	\hline
	
\end{tabular}

}

\caption[loftitle]{Overview of publicly available egocentric video datasets. Row one deals with object recognition. Rows 2-5 deal with activity detection/recognition. Rows 6 and 7 deal with life logging video data.  \label{tab:datasets}}
\end{figure*}

Figure \ref{tab:datasets} gives a compact overview over all datasets from the work mentioned in section \ref{sec:work} that are publicly available. We briefly describe the data as well as what kind of labeling is provided and also list the URLs to websites that contain further explanations and download links.

Most authors try to establish their own dataset and consequently none of the datasets has taken over the role of a true benchmark dataset. An exception might be the ``Intel 42 Objects'' dataset for the task of egocentric object recognition, which has also been used by \cite{fathi2011learning} to test the performance of their motion-based foreground-background segmentation method. Further, the ``Activities of Daily Living'' dataset was used by \cite{Lu_2013_CVPR} to test their story-driven video summarization method. However, as this dataset was primarily collected for the task of activity recognition \cite{pirsiavash2012detecting}, a direct comparison between both works was not possible.

\section{Summary and Comparison}
\label{sec:summary}

In this section, we summarize the key aspects of the work that was introduced in the previous sections and draw comparisons where possible.

Ren and Philipose  \cite{ren2009egocentric} were the first to test standard recognition systems for the task of recognizing handled objects in egocentric video. They continued to find that foreground-background segmentation can successfully be done with optical flow based approaches and helps to improve the recognition results, as handled objects tend to be in the foreground \cite{ren2010figure}. Their segmentation method was improved by Fathi \textit{et al.}, \cite{fathi2011learning} who also were the first to consider multiple objects being manipulated as part of kitchen activities like making sandwiches. Fathi \textit{et al.} went on to experiment with various weakly supervised approaches to recognize such activities, including object co-occurrence and changes in object states \cite{fathi2011understanding, Fathi_2013_CVPR}. They are also the only group to experiment with the influence of gaze with respect to activity recognition \cite{fathi2012learning}. Pirisiavash and Ramanan  \cite{pirsiavash2012detecting} were successful at recognizing more versatile household activities. However, unlike the work of Fathi \textit{et al.}, their method is strongly supervised. Ryoo and Matthies started looking at interaction level activities such as shaking hands \cite{S._2013_CVPR}. They discovered that activities that contain a lot of ego-motion can be well described with optical flow based approaches. Kitani \textit{et al.} \cite{kitani2011fast} came to similar conclusions when looking at sport activities that also involve a lot of ego-motion.

In parallel, researchers started looking at egocentric video for life logging purposes. Doherty \textit{et al.} \cite{doherty2008investigating} were the first to investigate keyframe selection methods in egocentric video, finding that a lot of the established methods to segment video into coherent parts do not work well due to the continuous nature of the video data. Followup work by Lee \textit{et al.} as well as Lu and Grauman \cite{lee2012discovering, Lu_2013_CVPR} investigated import objects and people as features to build better methods for keyframe extraction and summarization of egocentric life logging video. In contrast, Aghazadeh \textit{et al.} looked at life logging video of one subject over multiple days and detected novel or out of the ordinary activities.

\section{Conclusion}
\label{sec:conclusion}

In the previous sections, we gave a broad overview regarding the different problems in the domain of egocentric video that have recently been addressed in the computer vision community. We showed that research could roughly be grouped into three categories: object recognition, activity and action detection, life logging video summarization. All work in this domain is at a very early stage: The first publications on egocentric object recognition \cite{ren2009egocentric} and action segmentation \cite{spriggs2009temporal} date back to the first (out of two) IEEE workshop on egocentric vision during CVPR 2009. Early work on egocentric video in life logging scenarios  only dates back to 2008 \cite{doherty2008investigating}. As one result of this, almost all publications introduce their own, novel data sets while working with other authors' data remains the exception. Consequently, no dominant benchmark datasets have emerged so far like they have in other computer vision areas such as general object recognition.

Despite the novel nature of the egocentric vision domain, we can see some trends that span across all research categories: Egocentric video is all about objects. In first person videos, objects of interest tend to be naturally centered and at a large scale while being subject to relatively little occlusion, which makes egocentric video very convenient for object detection and classification. Additionally, optical flow based methods seem to work very well for the task of segmenting foreground objects (that are manipulated by hands) from background noise and are used in almost all recent publications to improve recognition results. This object-centered idea expands to action and activity recognition. Traditional work in this area (with video from third person cameras) usually involves approaches that use body configurations and movements as main features and try to detect, for instance, sport activities. In contrast, activities that are interesting from an egocentric perspective almost always involve objects that are being manipulated, while body movements are of little help. Consequently, almost all the work on activity recognition presented in section \ref{sec:work} used object detection in some way. Analogously, a lot of the work on life logging summarization uses interacted objects as cues for interesting or representative frames, resulting in better keyframes and summarizations than commonly used methods.

\bibliographystyle{IEEEtran}
{\footnotesize 
\bibliography{survey_cv}
}

\end{document}